\documentclass[10pt,twocolumn,letterpaper]{article}

\usepackage{amsmath, amsthm, amssymb, bbm, bm}
\newcommand\cl{\texttt{CL}\xspace}
\newcommand\corloc{\texttt{CorLoc}\xspace}

\newcommand\ytovone{\texttt{YTOv1}\xspace}
\newcommand\ytovonebold{{\fontfamily{lmtt}\fontseries{b}\selectfont YTOv1}\xspace}
\newcommand\ytovtwodtwo{\texttt{YTOv2.2}\xspace}
\newcommand\ytovtwodtwobold{{\fontfamily{lmtt}\fontseries{b}\selectfont YTOv2.2}\xspace}
\newcommand\deit{\texttt{DeiT}\xspace}
\newcommand\clipseg{\texttt{CLIPSeg}\xspace}
\newcommand\clip{\texttt{CLIP}\xspace}

\newcommand\cls{\texttt{class}\xspace}

\newcommand{\trm}[1]{\mathrm{#1}}

\newcommand{\softmax}{\trm{softmax}}

\newcommand\tableplus[1]{\textcolor{darkergreen}{#1}}
\usepackage{times}
\usepackage{hhline}

\usepackage{cvpr}              % To produce the CAMERA-READY version
\usepackage[dvipsnames]{xcolor}
\definecolor{cvprblue}{rgb}{0.21,0.49,0.74}
\usepackage[pagebackref,breaklinks,colorlinks,citecolor=cvprblue]{hyperref}

\title{Leveraging Transformers for Weakly Supervised Object Localization\\ in Unconstrained Videos}

\author{Shakeeb~Murtaza$^{1}$,
  ~Marco~Pedersoli$^{1}$,
  ~Aydin~Sarraf$^{2}$, and
  ~Eric~Granger$^{1}$\\
 $^1$ LIVIA, Dept. of Systems Engineering, ETS Montreal, Canada\\
$^2$  Ericsson, Global AI Accelerator, Montreal, Canada\\
{\tt\small shakeeb.murtaza.1@ens.etsmtl.ca,}
}

\usepackage{array}
\usepackage{xcolor}
\usepackage{colortbl}
 \usepackage{multirow} 
 
\definecolor{darkergreen}{RGB}{21, 152, 56}
\definecolor{red2}{RGB}{252, 54, 65}
\definecolor{Gray}{gray}{0.85}
\usepackage{epsfig}
\usepackage{graphicx}
\newcolumntype{g}{>{\columncolor{Gray}}c}

\begin{document}
\maketitle

\begin{abstract}
Weakly-Supervised Video Object Localization (WSVOL) involves localizing an object in videos using only video-level labels, also referred to as tags. State-of-the-art WSVOL methods like Temporal CAM (TCAM) rely on class activation mapping (CAM) and typically require a pre-trained CNN classifier. However, their localization accuracy is affected by their tendency to minimize the mutual information between different instances of a class and exploit temporal information during training for downstream tasks, e.g., detection and tracking. In the absence of bounding box annotation, it is challenging to exploit precise information about objects from temporal cues because the model struggles to locate objects over time. To address these issues, a novel method called \textit{transformer based CAM for videos (TrCAM-V)}, is proposed for WSVOL. It consists of a \deit backbone with two heads for classification and localization. The classification head is trained using standard classification loss (\cl), while the localization head is trained using pseudo-labels that are extracted using a pre-trained \clip model. From these pseudo-labels, the high and low activation values are considered to be foreground and background regions, respectively. Our TrCAM-V method allows training a localization network by sampling pseudo-pixels on the fly from these regions. Additionally, a conditional random field (CRF) loss is employed to align the object boundaries with the foreground map. During inference, the model can process individual frames for real-time localization applications. Extensive experiments\footnote{
Code: \href{https://github.com/shakeebmurtaza/TrCAM/}{https://github.com/shakeebmurtaza/TrCAM/}} on challenging YouTube-Objects unconstrained video datasets show that our TrCAM-V method achieves new state-of-the-art performance in terms of classification and localization accuracy. 
\end{abstract}

%%%%%%%%%%%%%%%%%%%%%%
\section{Introduction}

The rapid development of video-sharing platforms has led to the availability of extensive video data~\cite{ShaoCSJXYZX22,tang2013discriminative}, escalating the need for automated tools that analyze video content. Video object localization plays a crucial role in understanding this content. It also helps in enhancing the performance of our model in different downstream tasks, such as  
video object detection~\cite{Chen0HW20,HanWCQ20,ShaoCSJXYZX22}, visual object tracking~\cite{BergmannML19,LuoXMZLK21}, video summarization~\cite{ZhangHJYC17} and event detection~\cite{ChangYLZH16}.

Most videos are captured in wild and unconstrained environments, exhibiting different properties due to different factors such as differences in camera lenses, multiple viewpoints, decoding distortions, moving objects, and editing effects. Leveraging these videos for training our model to perform different downstream tasks requires bounding boxes or pixel-level labels by human annotators. In contrast to images, labeling videos is a very arduous and expensive task, as each video contains a large number of frames. To reduce the annotation cost, videos are weakly labeled \cite{JerripothulaCY16,tsai2016semantic} by identifying a global tag/class for each video. These global tags represent the object of interest in the video even though some frames may not contain the object of interest, leading to inconsistent labels for different frames. Moreover, these labels don't represent spatial-temporal information about a particular object across different frames. This results in noisy or corrupted frame-level labels, as labels are attributed to an entire video, even though only a subset of its frames might contain the object of interest. 

Using these global noisy labels for object localization can reduce the dependency on bounding box annotations, making this task more challenging. Different techniques have been proposed for weakly supervised video object localization (WSVOL)~\cite{joulin2014,jun2016pod,Kwak2015,prest2012learning,RochanRBW16,zhang2020spftn} and weakly supervised video object segmentation~\cite{tcamsbelharbi2023,Croitoru2019,FuXZL14,Halle2017,LiuTSRCB14,tsai2016semantic,Tokmakov2016,umer2021efficient,YanXCC17,ZhangJS14}. Nevertheless, these techniques necessitate post-processing, rendering them infeasible for object localization in unconstrained videos within real-world scenarios.

State-of-the-art WSVOL methods closely adhere to standard protocols~\cite{HartmannGHTKMVERS12,Kwak2015,prest2012learning,tang2013discriminative,Tokmakov2016,XuXC12,YanXCC17,zhang2020spftn}. They first generate object proposals using visual cues (e.g., motion) which are then employed to identify the relevant object using different post-processing techniques. Despite their outstanding performance, these methods exhibit several limitations. Typically, these methods involve multiple steps for WSVOL that restrain us from training these models in an end-to-end fashion, thus rendering these models susceptible to sub-optimal solutions. They also necessitate a separate model for each class as they are trained to localize a particular object. These limitations deem these models ill-suited for deployment in real-world environments and restrict their scalability to a small number of classes due to various constraints (e.g., resources and inference time). Most of these methods cannot employ class-level labels for extracting initial proposals, so they cannot semantically align different object parts belonging to an object of a particular class/tag. Moreover, these methods often utilize motion cues (e.g., optical flow) to localize an object of interest, which makes them susceptible to the same alignment problem as they often ignore the semantics of a particular object. Also, relying solely on motion cues hinders the network performance as they contain noisy information in unconstrained videos due to various external factors such as camera movements.

In a recent study, a discriminative multi-class deep learning model (DL) for WSVOL has been proposed~\cite{tcamsbelharbi2023}. This method leverages class activation maps (CAMs) for WSVOL, which have proven effective for weakly supervised object localization (WSOL) in static images~\cite{choe2020evaluating, belharbi2022fcam, murtaza2022dips, murtaza2023dips, murtaza2022dipssypo, rony2023deep}. In WSOL, the model is trained using image-level labels to localize an object of interest corresponding to the underlying class. These CAMs highlight the areas that strongly contribute to predicting a particular class. Nevertheless, these methods are not inherently designed to utilize the temporal information in videos for WSVOL. To deal with this issue, recent studies~\cite{tcamsbelharbi2023} propose an approach to leverage spatio-temporal information by harvesting CAMs using the LayerCAM~\cite{zhou2016learning} method. 
While this approach harnesses the spatiotemporal relationships within videos, it remains susceptible to the accumulation of inaccurate activations arising due to unconstrained object motion. This prevents the model from forming dependencies between object parts, thereby hindering its ability to localize various parts of the object, as depicted in the results section of their study.

To address these issues, we propose a transformer-based class activation mapping for videos (TrCAM-V) to localize a particular object. Unlike other methods, TrCAM-V requires only video-level annotations for training and is independent of additional assumptions, such as motion and temporal cues. This is inspired by WSOL methods, which employ pseudo-labels to train models for WSOL tasks within static images~\cite{murtaza2022dips, belharbi2022fcam, wei2021shallowspol}. They utilize pseudo-labels and class-level labels to train the localization and classification head in an end-to-end fashion. Moreover, these approaches scale effectively with a large number of classes, making them applicable to a wide range of applications.

This paper aims to explore the transformer models for WSVOL tasks by levering pseudo-label from a Contrastive Language-Image Pretraining (\clip) model.  Our proposed method, called TrCAM-V, consists of a transformer encoder (adhering to the design framework of \deit~\cite{touvron2021training}) with two heads -- one head for classification and the other for localization. The localization head produces a map with the same resolution as the input image, which is trained using pseudo-labels. Furthermore, the pseudo-labels for each frame are extracted from \clip~\cite{radford2021learning} model using GradCAM presented in \cite{lin2023clip}. This model accepts an image with a text prompt (class label) to produce a pixel-level pseudo-label (activation map). In line with standard protocols \citep{durand2017wildcat,zhou2016learning}, the strong and weak activations are deemed as FG and BG regions, respectively. Moreover, at each stochastic gradient descent (SGD) step~\cite{murtaza2022dips}, foreground (FG) and background (BG) pseudo-pixels are selected to build pseudo-labels~\cite{murtaza2022dips} to train localization head. This random sampling enables the network to explore FG/BG regions and fosters the emergence of activation values over different object parts in the localization map. We also employ conditional random field (CRF) loss to align the object boundaries with the boundaries of the localization map by leveraging statistical properties of an image, such as pixel color and proximity among pixels. Moreover, our model does not require exploiting temporal dependencies during either training or inference. This approach is more suitable for real-time applications compared to other state-of-the-art WSVOL methods, as TCAM does not need to process an entire video to localize an object within a particular frame. 

The main contributions of this paper are summarized as follows. 

\noindent \textbf{(1)} A novel transformer-based CAM (TrCAM-V) method is proposed for WSVOL tasks. It is comprised of two heads -- a classification head trained with class labels, and a localization head trained with pseudo pixels, which are obtained from a pre-trained \clip model.

\noindent \textbf{(2)} Unlike previous WSVOL methods~\cite{tcamsbelharbi2023}, TrCAM-V does not necessitate a pre-trained classifier to harvest activation maps and does not require temporal information during training.

\noindent \textbf{(3)} Following the WSVOL experimental protocol in \cite{tcamsbelharbi2023}, our empirical results  shows that our method achieves state-of-the-art performance the challenging YouTube-Object v1.0 (\ytovone)~\cite{prest2012learning} and v2.2 (\ytovtwodtwo)~\cite{KalogeitonFS16} datasets.

%%%%%%%%%%%%%%%%%%%%%%%%%%%%
\section{Related Work}

%%%%%%%%
\noindent \textbf{(a) Weakly Supervised Object Localization in Images.} The baseline method for WSOL in still images is class activation mapping (CAM). This technique generates the localization map by weighted aggregation of different activations map from the penultimate layer of a convolutional neural network (CNN)~\cite{zhou2016learning}. The weight coefficient for each activation map is determined by employing global average pooling (GAP) on these activations.
To improve the localization performance, different pooling layers have been introduced such as WILDCAT ~\citep{durand2017wildcat,durand2016weldon}, Peak Response Maps~\citep{ZhouZYQJ18PRM}, Log-Sum-Exp ~\citep{pinheiro2015image,sun2016pronet}, and multi-instance learning pooling ~\cite{ilse2018attention}. Despite the success of these methods, they tend to produce blobby maps covering only discriminative areas. This problem arises because these methods seek to minimize the mutual information between different instances of the same class, resulting in models that only highlight discriminate regions of a particular object~\cite{choe2020evaluating}. To deal with this issue, different methods have been proposed to expand attention maps beyond discriminative regions~\cite{belharbi2020minmaxuncer,ChoeS19,LiWPE018CVPR,MaiYL20eil,SinghL17,wei2017object,YunHCOYC19,ZhangWF0H18,zhu2017soft}. Additionally, different transformer based methods have been proposed to improve the effective receptive field and cover all object parts~\cite{,chen2022lctr,li2022caft,gupta2022vitol,bai2022weakly, su2022re,meng2022adversarial,gao2021tscam}. 

Instead of just minimizing the classification loss, different methods proposed to employ pseudo-labels for directly minimizing loss over generated maps to generate robust localization maps.~\cite{murtaza2022dipssypo,belharbi2022fcam,negevsbelharbi2022,MeethalPBG20icprcstn,murtaza2022dips,wei2021shallowspol,ZhangCW20rethink,ZhangWKYH18}. For instance, \cite{murtaza2022dipssypo, murtaza2022dips}~efficiently sample different FG and BG pixels at each stochastic gradient descent step to reduce the effect of cluttered BG.  These methods rely solely on a forward pass to compute localization maps. However, different methods have been proposed to compute maps using both forward and backward passes~\cite{cao2015look,zhang2018top,SelvarajuCDVPB17iccvgradcam,ChattopadhyaySH18wacvgradcampp,fu2020axiom,JiangZHCW21layercam,Adebayo2018,Kindermans2019,desai2020ablation,WangWDYZDMH20scorecam,naidu2020sscam,naidu2020iscam}. For instance,~\cite{ChattopadhyaySH18wacvgradcampp,fu2020axiom,JiangZHCW21layercam} rely on gradient information while~\cite{Adebayo2018,Kindermans2019,desai2020ablation,WangWDYZDMH20scorecam,naidu2020sscam,naidu2020iscam} employees confident aggregation for avoiding gradient saturation. Similarly,~\cite{ChattopadhyaySH18wacvgradcampp} and~\cite{fu2020axiom} employee feedback layer and  Excitation-backprop to improve localization performance of the network.

%%%
\noindent \textbf{(b) Weakly Supervised Video Object Segmentation.} Several methods have been proposed for video segmentation, each requiring various post-processing steps to produce final activation maps. Most of these methods process single videos or clusters of videos to localize a prominent object without relying on discriminative information.

Baseline methods for WSVOL~\cite{HartmannGHTKMVERS12,tang2013discriminative,XuXC12,YanXCC17} begin by extracting sptial-temporal segmentation either using pre-trained detectors~\cite{zhang2015semantic} or some unsupervised methods~\cite{BanicaAIS13,XuXC12}. These methods employ different graph-based techniques such as GrabCut~\citep{RotherKB04} and conditional random fields (CRF)~\cite{krahenbuhl2011efficient,tang2018regularized}. The extracted segments are then used to localize an object by imposing motion cues and the visual appearance of a particular object. Similarly, \cite{LiuTSRCB14}~proposed a method for multi-class segmentation by employing nearest neighbor-based label transfer methods between videos belonging to the same subclass. Initially, spatial-temporal supervoxels~\cite{XuXC12} are identified that are projected into a high-dimensional space using different cues e.g., color, texture, and motion. Using these supervoxels, a graph is constructed to enforce label smoothness among spatial-temporally contiguous supervoxels in a particular video, as well as among supervoxels exhibiting similar visual appearances across various videos. Additionally, M-CNN~\cite{Tokmakov2016} employs a CNN network to estimate FG regions using motion cues and the Gaussian mixture model. These FG regions are then integrated using a fully convolutional network fine-tuned on a few images.

Several methods have also been proposed for co-segmentation, where a model is trained to identify similar objects presented in different images. A common approach is to utilize inter and intra-video visual and motion cues to discover common segments by modeling the relations between different segments using graph-based techniques~\cite{ChenCC12,FuXZL14,tsai2016semantic,ZhangJS14}. For instance,~\cite{ZhangJS14} employed a regulated maximum weight clique to sample object proposal for co-segmentation. Similarly, \cite{fu2015object}~ employs a method for multiple-object segmentation in a scene by using intra-video coherence of different object parts, as well as the consistency of the FG objects across various videos. Moreover, in~\cite{tsai2016semantic} the authors extracted object tracklets for each video using a pre-trained FCN and linked them to the object category using a graph. Using this graph, relationships between different tracklets were formulated using various cues (e.g., shape, motion, appearance) to discover the prominent objects in different videos.

The methods described above optimize using global class labels for each video. However, different methods have been proposed to learn object segmentation without relying on video level labels~\citep{Croitoru2019,Halle2017,papazoglou2013fast,umer2021efficient}. These methods first estimate potential FG regions using different cues (motion, appearance) and then use these regions to produce a segmentation map. For example,~\citep{Croitoru2019} employed a DL model to process initial FG regions as pseudo-labels to train CNN for segmentation. 

\begin{figure*}[!t]
\centering
\includegraphics[width=0.8\textwidth, trim={5 5 5 5},clip]{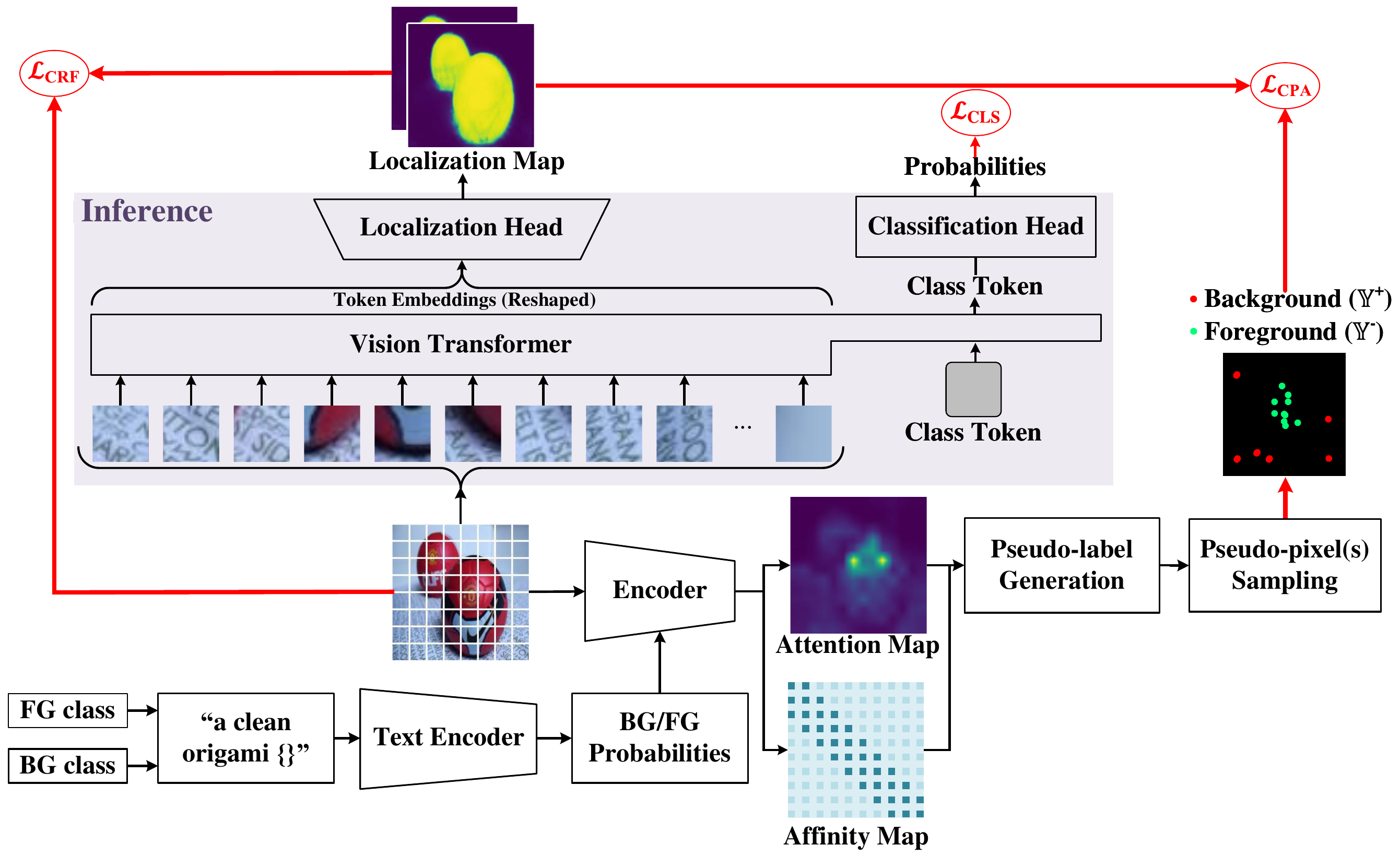}
\caption{Illustration of the proposed TrCAM-V training architecture. It consists of \deit backbone a classification and a localization head that are trained using class labels and pseudo labels, respectively. A pre-trained \clip model is employed to generate pseudo-labels by utilizing a sharpness-based prompt along with the input image, as suggested in \cite{lin2023clip}. These pseudo-labels are then used to sample pseudo-pixels for training the localization head. For inference, we only retain \deit with both heads.}\label{fig:overview}
\end{figure*}

%%%
\noindent \textbf{(c) Weakly Supervised Video Object Localization.}
Despite the success of deep learning methods for WSOL in still images, limited methods have been proposed for WSVOL~\citep{tcamsbelharbi2023,belharbi2023colo,jun2016pod,Kwak2015,prest2012learning,RochanRBW16,zhang2020spftn}. Most of these models typically initialize and refine prominent proposals while incorporating spatiotemporal consistency constraints by exploiting information from visual appearance and motion cues of an object. Some methods have employed proposals as supervision to train localizers~\citep{prest2012learning,zhang2020spftn}, whereas others have leaned towards segmentation-based approaches~\citep{RochanRBW16}, further refining the results using GrabCut~\citep{RotherKB04}. Moreover, these methods often only select one or more videos of a particular class to optimize an underlying model.

Similarly, methods have been proposed to localize a prominent object in a video by identifying similar FG regions based on visual appearance and motion cues~\cite {jun2016pod,RochanRBW16,Kwak2015}. For instance, \cite{RochanRBW16}~generate bounding box proposals within a video and retain relevant ones to build an object appearance detection model by enforcing temporal appearance consistency using maximum a posteriori inference. Similarly, \cite{Kwak2015}~learn to localize an object by discovering similar objects in different videos and tracking prominent regions in individual videos. \cite{ManenGG13}~employed region proposals along with appearance and motion consistency to discover FG objects~\cite{BroxM10} thereby maintaining temporal relationships between consecutive frames. In addition to conventional methods, deep learning (DL) models have been proposed for WSVOL. For instance, ~\citep{zhang2020spftn} proposed SPFTN to jointly learn object segment and localization tasks by employing an optical flow technique~\citep{LeeKG11}. Recently~\cite{tcamsbelharbi2023, belharbi2023colo} proposed an approach for WSVOL that exploits mutual information between different consecutive frames. However, these methods require temporal information during training and also rely on pre-trained classifiers to harvest pseudo-labels and low-level representations that hinder their localization abilities as they tend to minimize the mutual information between instances of the same class. Compared to these methods, TrCAM-V does not require temporal information during training, yet is capable of achieving state-of-the-art performance because it does not rely on discriminative pseudo-labels, harvested from a pre-trained classifier, for the training of the localization head.

%%%%%%%%%%%%%%%%%%%%%%%
\section{Proposed Approach}

\noindent \textbf{Notation.} Consider a training set $\mathbb{T} = \{\mathrm{v}_i, \mathrm{y}_i\}$ where each $\mathrm{v}_i=\{x_j\in\mathbb{R}^{H\times W\times 3}\}_{j=1}^f$ represent an input video composed of $f$ frames and  each $\mathrm{y}_i\in\{1,\dots,C\}$ represent the global class label $i$-th video, where $i = 1, 2, \dots, n$ and $C$ is the number of classes in our dataset. It's further presumed that each frame within a video encompasses a particular object corresponding to the underlying class $\mathrm{y}_i$. Furthermore, our model is capable of accepting a frame $f_i$ and predicting class category $y_i$ along with a localization map $\hat{M}$ encompassing a corresponding object. 
%%%%%%%%%%
\subsection{Background on Transformers}

\noindent \textbf{Vision Transformer (ViTs)} have recently 
achieved state-of-the art accuracy in different tasks such as classification~\cite{dosovitskiy2020image, touvron2021training} and WSOL \cite{gao2021tscam,murtaza2022dips} tasks. A ViT is composed of $N$ cascaded encoder blocks that capture long-range relationships between different object parts within a scene. Each block consists of different layers, including a multi-headed attention layer, followed by a fully connected layer. To pass an image through the transformer, a frame $f_i$ is divided into $M$ patches of size $(H/S)\times (W/S)$. Subsequently, positional embeddings and a class (\cls) token are added to the $P$ patches and projected into a fixed embedding dimension of size $d$. The computed embeddings are then passed through the $N$ cascaded encoder blocks of the transformer. From the output of the last encoder layer, the \cls token is extracted and passed through a fully connected layer to produce classification scores for the relevant classes.

\noindent \textbf{Self-Supervised Transformers (SSTs)} can learn object representations within an image without relying on explicit supervision. In this context, self-distillation with no labels (DINO) has been proposed to learn meaningful representation from data without supervision~\cite{caron2021emerging}. To learn a representation in an unsupervised fashion, the author employed a student and a teacher network. The objective is to match their probability distributions corresponding to two different views of an input image. The output of these networks is normalized using temperature $\softmax$. Moreover, after each optimization step, the parameters of the student are transferred to the teacher network using an exponential moving average.

%%%%%%%%
\subsection{Transformer-Based CAM for Videos}
The SST~\cite{caron2021emerging} can identify objects in a frame without explicit supervision. However, they tend to decompose different object parts each corresponding to different attention maps as they are trained without explicit supervision. These attention maps highlight different object parts based on their attention coefficients learned during network training. For extracting relevant object parts and suppressing BG regions from these maps, we propose TrCAM-V. It decodes the patch embeddings and \cls tokens to produce a localization map that is optimized using pseudo-labels derived from the pre-trained \clip model. Furthermore, these pseudo-labels allow us to directly optimize the loss over generated pseudo-labels instead of just minimizing the mutual information between different instances of a particular class.

More specifically, our model consist of a transformer backbone $T$ with two heads; \textbf{(i)} classification head $F$ trained using global video labels via standard cross-entropy, ${\min_{\bm{\theta}}- \log(\mbox{Pr}(y | f_i))}$. It accept \cls token for an input video frame $f_i$ from transformer backbone and produce classification probabilities for each class $F(T(f_i))=\mbox{Pr}(c|f_i)$ where $F(T(f_i))\in[0,1]_c$. \textbf{(ii)} A localization head $G$ that accepts patch embeddings and \cls tokens from the last layer of the transformer to produce a high-resolution localization map $M$. It consists of two channels $M_0$ and $M_1$, which represent the BG and FG maps, respectively. The localization head is trained using pseudo-labels extracted from a pre-trained \clip model (PCM). To extract the pseudo-label, TrCAM-V first obtains a saliency map from PC, after which it determines the FG and BG regions using Otsu's threshold~\cite{otsuthresh}. The densest region among the different regions extracted by the Otsu method is used to sample FG and BG regions. Given that these sampled regions may be noisy and uncertain regarding our concerned object, we refrain from directly using these maps to minimize the loss over the generated map. Instead, we stochastically sample a few FG and BG pixels~\cite{feng2017discriminative} and utilize them to train our model using partial cross-entropy. This allows the network to explore relevant object parts, resulting in a more robust localization map. This sampling process can also be viewed as a \textit{fill-in-the-gap} approach. It enables our localization head to explore similar-looking object parts while considering the statistical properties of the object to generate a robust localization map. Moreover, in contrast to baseline methods~\cite{tcamsbelharbi2023,belharbi2023colo}, we do not take into account the temporal information for producing pseudo-labels. 

\noindent \textbf{Initial pseudo-label generation.}
Following~\cite{lin2023clip}, we employ GradCAM to harvest maps from \clip for FG maps while suppressing regions belonging to BG categories. Initially, attention map regions are identified via a text-driven approach, leveraging a sharpness-based prompt as opposed to solely relying on class categories. To extract weights for generating GradCAM, a softmax function is applied to the computed attention between text and image embeddings. To compute the final activation maps, class-aware attention-based affinity (CAA) maps are extracted and fused with initial CAMs, as depicted in Fig.\ref{fig:overview}. These harvested maps can be viewed as pseudo-labels that are used to harvest pseudo-pixels.

\noindent \textbf{Selection of FG/BG pixels.} To train the localization head $G$, we leverage pseudo pixels from saliency maps $S$ obtained from pre-trained \clip model. We produced a separate pseudo-label for each frame $f_i$, without considering temporal information into account. In each saliency map, weak and strong activations are likely to represent the BG and FG regions, respectively~\cite{tcamsbelharbi2023,negevsbelharbi2022,belharbi2022fcam,durand2017wildcat,murtaza2022dips,zhou2016learning}. Relying on this assumption, we can estimate BG ${\mathbb{C}^-}$ and FG regions ${\mathbb{C}^+}$ using Otsu threshold~\cite{otsuthresh} as follows,

\begin{equation}
    \label{eq:sets}
    \mathbb{M}^+_t = \mathcal{O}^+(\bm{M}), \quad \mathbb{M}^-_t = \mathcal{O}^-(\bm{M})
\end{equation}
% \\
Here, operation $\mathcal{O}^+$ is employed to select top $n\%$ pixels from the most dense connected area in activation map $\bm{M}$. It selects pixels that have a magnitude larger than the Otsu threshold. Conversely, the operation $\mathcal{O}^-$ is used to select the bottom $-n\%$ pixels from $\bm{M}$, which are ordered from the lowest to highest activation value. These selected regions are uncertain about object location and may contain incorrect labels. Considering the potential uncertainty, a few BG and FG pixels are sampled probabilistically to construct our pseudo-labels, as follows:
\begin{equation}
    \label{eq:sset}
    \bm{M}^{\prime} = \mathcal{P}(\mathbb{M}^+)\; \cup \; \mathcal{P}(\mathbb{M}^-) \;, 
\end{equation}
where $\mathcal{P}$ represents the sampling process of a few pixels from selected FG and BG regions. To generate the final pseudo-label, the values of the selected FG and BG pixels are replaced with 1's and 0's, respectively.

\noindent \textbf{Training loss.} Cross-entropy is employed to train our classification head. It relies on \deit features for producing class category $\mbox{Pr}(c | \bm{f_i})$ as follows: 
\begin{equation}
    \label{eq:cl-loss}
    \min_{\bm{\theta}_c} \;  - \log(\mbox{Pr}(y | \bm{f_i}))\;.
\end{equation}
For the localization head, we employ a loss function consisting of three terms: \\ 
(i) The pixel alignment loss ($\mathcal{L}_{PAL}$) for learning FG/BG regions\footnote{A similar approach for exploiting few pixels to localize a concerned object is also employed in \cite{feng2017discriminative}.}. It aims to align the output map $\hat{M}$ with the selected pixels within $\mathcal{P}(\mathbb{M}^+)\; \cup \; \mathcal{P}(\mathbb{M}^-)$ via a partial cross-entropy loss denoted as $\mathcal{L}_{CPA}(\mathcal{P}(\mathbb{M}^+)\; \cup \; \mathcal{P}(\mathbb{M}^-), M^{r})$, where $r$ represents the pseudo-pixels: 
\begin{equation}\label{eq:pl}
\mathcal{L}_{PAL} = -(1-\hat{M})\log(\mathbb{M}^+)-\hat{M}(\log(\mathbb{M}^-))
\end{equation}
(ii) The absolute size loss ($\mathcal{L}_{ASL}$) for localizing all object parts belonging to FG/BG regions, as defined by \cite{belharbi2022fcam}. It is formulated as a constraint for pushing FG and BG regions away and optimized using the log-barrier method \cite{boyd2004convex}.\\ 
(iii) The conditional random field (CRF) loss is employed to align the localization map with the boundaries of the object by utilizing colour similarity and proximity of nearby pixels as $\mathcal{L}_{CRF}(\bm{S}, \bm{f_i}) =\sum_{r \in \{0, 1\}} {\bm{S}^r}^{\top} \; \bm{W} \; (\bm{1} - \bm{S}^r)$. $\bm{W}$ is the affinity matrix to capture colour similarity and pixels' proximity between pixels of frame $\bm{f_i}$ and $\bm{W}$. The overall loss is defined as: 
\begin{equation}\label{loss:confidence}
        \mathcal{L} = \min_{\theta} \lambda_{PAL}\mathcal{L}_{PAL} + \lambda_{ASL} \mathcal{L}_{ASL}  + \lambda_{CRF} \mathcal{L}_{CRF} \ , 
\end{equation}
where $\lambda_{CLS}$ and $\lambda_{CPA}$ are hyperparameters that lie in the interval $\left[0, 1\right]$. $2e^{-9}$ and $\lambda_{CRF}$ is set to $2e^{-9}$ \cite{tang2018regularized}.

\noindent \textbf{Inference.} During the inference phase, we discard \clip and its associated components utilized during training to harvest pseudo-labels. We only retain \deit backbone $T$ along with the classification and inference head. The backbone network $T$ inputs a frame $f_i$ and predicts a class label $y_i$ and the corresponding localization map $\hat{M}$.

%%%%%%%%%%%%%%%%%%%%%
\section{Results and Discussion}

\begin{table*}[h]
\begin{center}
\resizebox{\linewidth}{!}{
\begin{tabular}{|c|l|*{10}{c|}|g|c|}
\hline
\textbf{Dataset} & \textbf{Method \emph{(venue)}} & \textbf{Aero} & \textbf{Bird} & \textbf{Boat} & \textbf{Car} & \textbf{Cat} & \textbf{Cow} & \textbf{Dog} & \textbf{Horse} & \textbf{Mbike} & \textbf{Train} & \textbf{Avg} %& \textbf{Time/Frame} 
\\
\hhline{-----------||---}
\noalign{\vspace{\doublerulesep}}
\hhline{-----------||---}
 &\cite{prest2012learning} {\small \emph{(cvpr,2012)}} & 51.7 & 17.5 & 34.4 & 34.7 & 22.3 & 17.9 & 13.5 & 26.7 & 41.2 & 25.0 & 28.5 %& N/A   
 \\
 &\cite{papazoglou2013fast} {\small \emph{(iccv,2013)}} & 65.4 & 67.3 & 38.9 & 65.2 & 46.3 & 40.2 & 65.3 & 48.4 & 39.0 & 25.0 & 50.1 %& 4s  
 \\
 &\cite{joulin2014} {\small \emph{(eccv,2014)}} & 25.1 & 31.2 & 27.8 & 38.5 & 41.2 & 28.4 & 33.9 & 35.6 & 23.1 & 25.0 & 31.0 %& N/A 
 \\
 &\cite{Kwak2015} {\small \emph{(iccv,2015)}}  & 56.5 & 66.4 & 58.0 & 76.8 & 39.9 & 69.3 & 50.4 & 56.3 & 53.0 & 31.0 & 55.7 %& N/A  
 \\
 &\cite{RochanRBW16} {\small \emph{(ivc,2016)}} & 60.8 & 54.6 & 34.7 & 57.4 & 19.2 & 42.1 & 35.8 & 30.4 & 11.7 & 11.4 & 35.8 %& N/A 
 \\
 &\cite{Tokmakov2016} {\small \emph{(eccv,2016)}} & 71.5 & 74.0 & 44.8 & 72.3 & 52.0 & 46.4 & 71.9 & 54.6 & 45.9 & 32.1 & 56.6 %& N/A  
 \\
 &POD~\cite{jun2016pod} {\small \emph{(cvpr,2016)}} & 64.3 & 63.2 & 73.3 & 68.9 & 44.4 & 62.5 & 71.4 & 52.3 & 78.6 & 23.1 & 60.2 %& N/A  
 \\
 &\cite{tsai2016semantic} {\small \emph{(eccv,2016)}} & 66.1 & 59.8 & 63.1 & 72.5 & 54.0 & 64.9 & 66.2 & 50.6 & 39.3 & 42.5 & 57.9 %& N/A  
 \\
 &\cite{Halle2017} {\small \emph{(iccv,2017)}} & 76.3 & 71.4 & 65.0 & 58.9 & 68.0 & 55.9 & 70.6 & 33.3 & 69.7 & 42.4 & 61.1 %& 0.35s  
 \\
&\cite{Croitoru2019} (LowRes-Net\textsubscript{iter1}) {\small \emph{(ijcv,2019)}} & 77.0 & 67.5 & 77.2 & 68.4 & 54.5 & 68.3 & 72.0 & 56.7 & 44.1 & 34.9 & 62.1 %& 0.02s  
\\
\multirow{6}{*}{\ytovone } & \cite{Croitoru2019} (LowRes-Net\textsubscript{iter2}) {\small \emph{(ijcv,2019)}} & 79.7 & 67.5 & 68.3 & 69.6 & 59.4 & 75.0 & 78.7 & 48.3 & 48.5 & 39.5 & 63.5 %& 0.02s
\\
&\cite{Croitoru2019} (DilateU-Net\textsubscript{iter2}) {\small \emph{(ijcv,2019)}} & 85.1 & 72.7 & 76.2 & 68.4 & 59.4 & 76.7 & 77.3 & 46.7 & 48.5 & 46.5 & 65.8 %& 0.02s 
\\
&\cite{Croitoru2019} (MultiSelect-Net\textsubscript{iter2}) {\small \emph{(ijcv,2019)}} & 84.7 & 72.7 & 78.2 & 69.6 & 60.4 & 80.0 & 78.7 & 51.7 & 50.0 & 46.5 & 67.3 %& 0.15s 
\\
&SPFTN (M)~\cite{zhang2020spftn} {\small \emph{(tpami,2020)}} & 66.4 & 73.8 & 63.3 & 83.4 & 54.5 & 58.9 & 61.3 & 45.4 & 55.5 & 30.1 & 59.3 %& N/A  
\\
&SPFTN (P)~\cite{zhang2020spftn} {\small \emph{(tpami,2020)}}& 97.3 & 27.8 & 81.1 & 65.1 & 56.6 & 72.5 & 59.5 & 81.8 & 79.4 & 22.1 & 64.3 %& N/A  
\\
&FPPVOS~\cite{umer2021efficient} {\small \emph{(optik,2021)}} & 77.0 & 72.3 & 64.7 & 67.4 & 79.2 & 58.3 & 74.7 & 45.2 & 80.4 & 42.6 & 65.8 %& 0.29s  
\\
% \hhline{--||---------||---}
\cline{2-13}
&CAM~\cite{zhou2016learning} {\small \emph{(cvpr,2016)}} & 75.0 & 55.5 & 43.2 & 69.7 & 33.3 & 52.4 & 32.4 & 74.2 & 14.8 & 50.0 & 50.1 %& 0.2ms  
\\
&GradCAM~\cite{SelvarajuCDVPB17iccvgradcam} {\small \emph{(iccv,2017)}} & 86.9 & 63.0 & 51.3 & 81.8 & 45.4 & 62.0 & 37.8 & 67.7 & 18.5 & 50.0 & 56.4 %& 27.8ms  
\\
&GradCAM++~\cite{ChattopadhyaySH18wacvgradcampp} {\small \emph{(wacv,2018)}} & 79.8 & 85.1 & 37.8 & 81.8 & 75.7 & 52.4 & 64.9 & 64.5 & 33.3 & 56.2 & 63.2 %& 28.0ms  
\\
&Smooth-GradCAM++~\cite{omeiza2019corr} {\small \emph{(corr,2019)}} & 78.6 & 59.2 & 56.7 & 60.6 & 42.4 & 61.9 & 56.7 & 64.5 & 40.7 & 50.0 & 57.1 %& 136.2ms  
\\
&XGradCAM~\cite{fu2020axiom} {\small \emph{(bmvc,2020)}} & 79.8 & 70.4 & 54.0 & 87.8 & 33.3 & 52.4 & 37.8 & 64.5 & 37.0 & 50.0 & 56.7 %& 14.2ms  
\\
&LayerCAM~\cite{JiangZHCW21layercam} {\small \emph{(ieee,2021)}} & 85.7 & 88.9 & 45.9 & 78.8 & 75.5 & 61.9 & 64.9 & 64.5 & 33.3 & 56.2 & 65.6 %& 17.9ms  
\\
&TCAM~\cite{tcamsbelharbi2023} {\small \emph{(wacv,2023)}} & 90.5 & 70.4 & 62.2 & 75.7 & 84.8 & 81.0 & 81.0 & 64.5 & 70.4 & 50.0 & 73.0 %& 18.5ms  
\\
% &CoLo-CAM {\small \emph{(corr,2023)}} & 90.4 & 74.0 & 91.8 & 87.8 & 78.7 & 80.9 & 89.1 & 74.1 & 85.1 & 68.7 & 82.1  \\
\cline{2-13}
&TrCAM-V with \clip (ours) & 91.7 & 77.8 & 91.9 & 94.0 & 84.8 & 81.0 & 83.8 & 77.4 & 77.8 & 87.5 & 84.8 %& 18.5ms  
\\
&TrCAM-V with \clipseg (ours) & 94.0 & 74.1 & 94.6 & 90.9 & 87.9 & 81.0 & 89.2 & 77.4 & 74.1 & 75.0 & 83.8 %& 18.5ms  
\\
\hhline{-----------||---}
\noalign{\vspace{1.5mm}}
\hhline{-----------||---}
&\cite{Halle2017}  {\small \emph{(iccv,2017)}}& 76.3 & 68.5 & 54.5 & 50.4 & 59.8 & 42.4 & 53.5 & 30.0 & 53.5 & 60.7 & 54.9 %& 0.35s   
\\
&\cite{Croitoru2019} (LowRes-Net\textsubscript{iter1}) {\small \emph{(ijcv,2019)}} & 75.7 & 56.0 & 52.7 & 57.3 & 46.9 & 57.0 & 48.9 & 44.0 & 27.2 & 56.2 & 52.2 %& 0.02s    
\\
&\cite{Croitoru2019} (LowRes-Net\textsubscript{iter2}) {\small \emph{(ijcv,2019)}} & 78.1 & 51.8 & 49.0 & 60.5 & 44.8 & 62.3 & 52.9 & 48.9 & 30.6 & 54.6 &  53.4 %& 0.02s    
\\
&\cite{Croitoru2019} (DilateU-Net\textsubscript{iter2}){\small \emph{(ijcv,2019)}} & 74.9 & 50.7 & 50.7 & 60.9 & 45.7 & 60.1 & 54.4 & 42.9 & 30.6 & 57.8 & 52.9 %& 0.02s   
\\
\multirow{6}{*}{\ytovtwodtwo } & \cite{Croitoru2019} (BasicU-Net\textsubscript{iter2}){\small \emph{(ijcv,2019)}} & 82.2 & 51.8 & 51.5 & 62.0 & 50.9 & 64.8 & 55.5 & 45.7 & 35.3 & 55.9 & 55.6 %& 0.02s  
\\
&\cite{Croitoru2019} (MultiSelect-Net\textsubscript{iter2}){\small \emph{(ijcv,2019)}} & 81.7 & 51.5 & 54.1 & 62.5 & 49.7 & 68.8 & 55.9 & 50.4 & 33.3 & 57.0 & 56.5 %& 0.15s  
\\
% \hhline{-----------||--}
\cline{2-13}
&CAM~\cite{zhou2016learning} {\small \emph{(cvpr,2016)}} & 52.3 & 66.4 & 25.0 & 66.4 & 39.7 & 87.8 & 34.7 & 53.6 & 45.4 & 43.7 & 51.5 %& 0.2ms  
\\
&GradCAM~\cite{SelvarajuCDVPB17iccvgradcam} {\small \emph{(iccv,2017)}} & 44.1 & 68.4 & 50.0 & 61.1 & 51.8 & 79.3 & 56.0 & 47.0 & 44.8 & 42.4 & 54.5 %& 27.8ms  
\\
&GradCAM++~\cite{ChattopadhyaySH18wacvgradcampp} {\small \emph{(wacv,2018)}} & 74.7 & 78.1 & 38.2 & 69.7 & 56.7 & 84.3 & 61.6 & 61.9 & 43.0 & 44.3 & 61.2 %& 28.0ms  
\\
&Smooth-GradCAM++~\cite{omeiza2019corr} {\small \emph{(corr,2019)}} & 74.1 & 83.2 & 38.2 & 64.2 & 49.6 & 82.1 & 57.3 & 52.0 & 51.1 & 42.4 & 59.5 %& 136.2ms  
\\
&XGradCAM~\cite{fu2020axiom} {\small \emph{(bmvc,2020)}} & 68.2 & 44.5 & 45.8 & 64.0 & 46.8 & 86.4 & 44.0 & 57.0 & 44.9 & 45.0 & 54.6 %& 14.2ms  
\\
&LayerCAM~\cite{JiangZHCW21layercam} {\small \emph{(ieee,2021)}} & 80.0 & 84.5 & 47.2 & 73.5 & 55.3 & 83.6 & 71.3 & 60.8 & 55.7 & 48.1 & 66.0 %& 17.9ms  
\\
&TCAM~\cite{tcamsbelharbi2023} {\small \emph{(wacv,2023)}} & 79.4 & 94.9 & 75.7 & 61.7 & 68.8 & 87.1 & 75.0 & 62.4 & 72.1 & 45.0 & 72.2 %& 18.5ms  
\\
% &CoLo-CAM {\small \emph{(corr,2023)}} & 82.9 & 92.2 & 85.4 & 67.7 & 80.1 & 85.7 & 79.2 & 67.4 & 72.7 & 58.2 & 77.1 \\
\cline{2-13}
&TrCAM-V with \clip (ours) & 87.6 & 91.6 & 90.3 & 74.1 & 78.7 & 79.2 & 76.2 & 66.9 & 60.0 & 62.0 & 76.7 %& 18.5ms 
\\
&TrCAM-V with \clipseg (ours) & 84.7 & 95.5 & 92.4 & 79.3 & 78.7 & 87.9 & 84.1 & 66.9 & 68.4 & 62.0 & 80.0 %& 18.5ms  
\\
\hline
\end{tabular}}
\end{center}
\caption{\corloc performance on \ytovone~\cite{prest2012learning} and \ytovtwodtwo~\cite{KalogeitonFS16} test sets. Results of related methods are borrowed from \cite{belharbi2023colo}.}
\label{tab:loc}
\end{table*}

%%%
\subsection{Experimental Methodology}

\noindent \textbf{Datasets.} To validate the performance of our model, we conducted extensive experiments on two challenging datasets comprised of unconstrained videos from YouTube\footnote{\href{https://www.youtube.com/}{https://www.youtube.com/}}; YouTube-Object
v1.0 (\ytovone)~\cite{prest2012learning} and YouTube-Object
v2.2 (\ytovtwodtwo)~\cite{KalogeitonFS16}. For training, only global video labels are available, which correspond to the prominent object in different frames of the video. Further details of these datasets are given below:

\noindent \ytovonebold consists of various videos collected from YouTube. Ten categories were considered to query different videos containing a particular object. Each class consists of a varying number of videos ranging from 9-24 frames with durations spanning from 0.5-3 minutes. The dataset comprises 155 videos that are segmented into short clips, referred to as shorts. After dividing the videos into clips, we ended up with 5507 shorts encompassing 571,089 frames. Within each short, a few frames have bounding box annotations corresponding to the underlying class. The annotated frames can be used as validation and test sets for hyper-parameter and model selection, respectively. Moreover, the authors reserved 128 videos for training and 27 for testing, which collectively encompass a total of 396 labeled bounding boxes. We reserved some frames from the training videos for a validation set. To build the validation set, we followed \cite{belharbi2022fcam} and randomly chose five videos per class, resulting in a total of 50 videos.

\noindent \ytovtwodtwobold is an extension of \ytovone and consist of 722,040 frames. Compared to \ytovone, this dataset has a large number of bounding box annotations for different categories. The dataset comprises 155 videos divided into 9 categories. In terms of dataset partitioning, the authors reserved 106 videos for training and 49 videos for testing. Following \cite{tcamsbelharbi2023}, we built our validation set by randomly selecting three videos per class from the training set. Compared to \ytovone, this dataset has a larger number of bounding box annotations; the test set comprises 1,781 frames with bounding box annotations, resulting in a total of 2,667 bounding boxes. The expanded test set and annotations render \ytovtwodtwo more challenging.

\noindent \textbf{Evaluation measures.}
We employ two metrics for measuring the performance of our model; \textbf{(i)} \corloc \cite{deselaers2012weakly} is used to measure the localization performance. \corloc indicates the proportion of predicted bounding having an intersection over union (IoU) greater than 50\%. (ii) Standard classification accuracy (\cl) is measured to evaluate the classification performance of our model. \cl is measured over frames that have bounding boxes annotations.

\noindent \textbf{Implementation details.} We follows the protocols of \cite{tcamsbelharbi2023} for all of our experiments. Specifically, we train our model for 100 epochs with a mini-batch size of 32. All images are resized to $256\times256$, then randomly cropped to a size of $224\times224$. After cropping, we augment the image by random horizontal flipping, followed by normalization. The weight $\lambda$ for CRF is set to $2e^{-9}$ as defined in~\cite{tang2018regularized}. For log-barrier optimization, we use the same hyperparameter value as suggested in~\cite{belharbi2019unimoconstraints,kervadec2019log,tcamsbelharbi2023}. We optimize our network using the Stochastic Gradient Descent (SGD) algorithm with learning rates ranging between 0.1 and 0.00001~\cite{murtaza2022dips}. Moreover, we use the~\deit backbone with pre-trained weights from SST~\cite{caron2021emerging}. To account for the large number of redundant frames per video, we randomly selected a few frames for each shot for each gradient descent step. This approach enabled us to train our model over a large number of video frames within a reasonable timeframe.
\begin{figure*}[!t]
\centering
\includegraphics[width=0.75\textwidth, trim={0 0 0 0},clip]{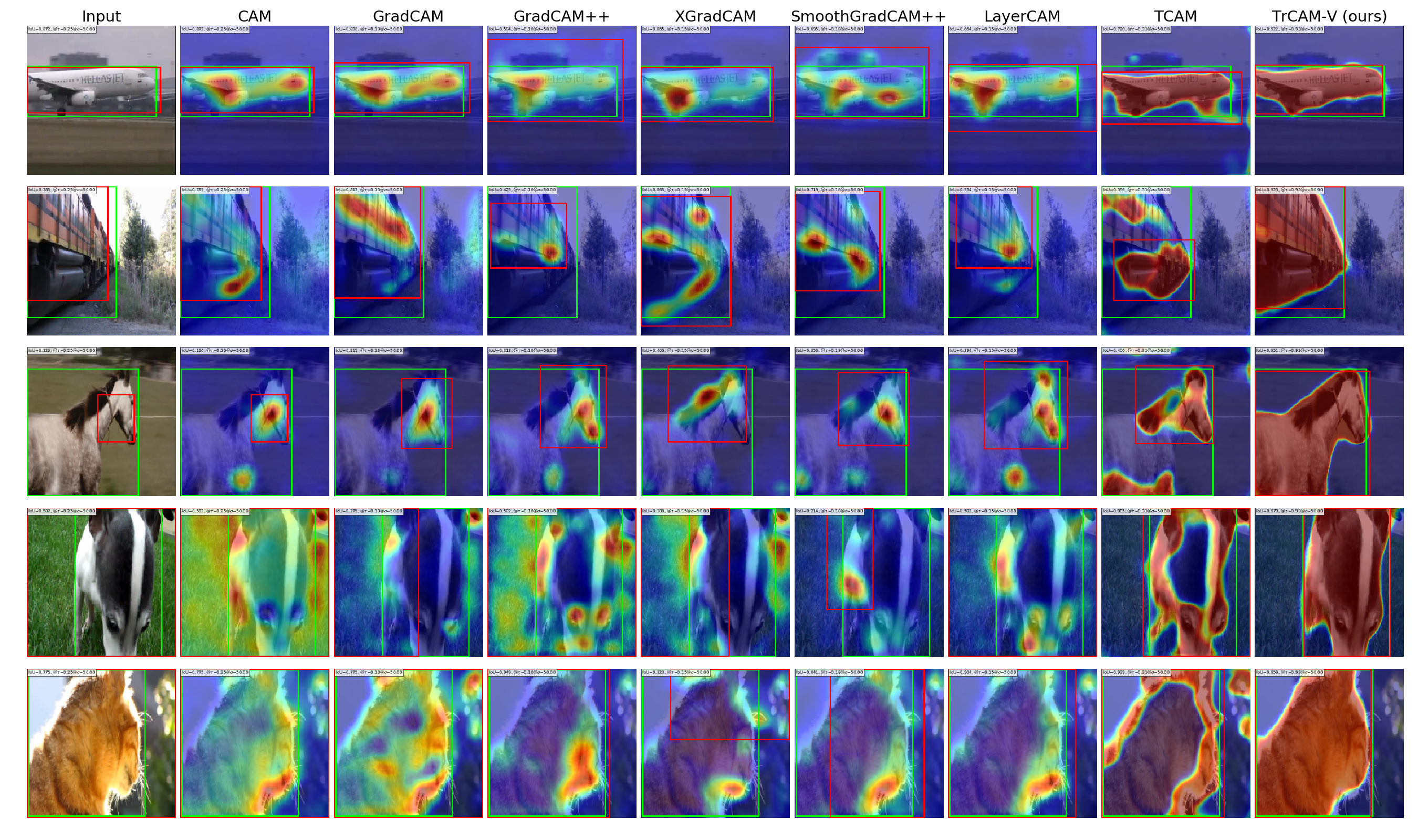}
\caption{Visualization of \ytovone frames. Here, {\color{red}red} and {\color{green}green box} indicate the predicated and annotated bounding- box.}\label{fig:results_yt_1}
\end{figure*}

{
\setlength{\tabcolsep}{3pt}
\renewcommand{\arraystretch}{1.1}
\begin{table}[ht!]
\centering
\resizebox{0.9\linewidth}{!}{%
\centering
\small
\begin{tabular}{lgg}
\toprule
\textbf{Methods} &  \ytovone & \ytovtwodtwo  \\
\hline
% \cline{1-1}\cline{3-3} \cline{5-5} 
CAM~\citep{zhou2016learning} {\small \emph{(cvpr,2016)}}  &                     85.3 & 73.9  \\
GradCAM~\citep{SelvarajuCDVPB17iccvgradcam} {\small \emph{(iccv,2017)}}  &                     85.3 & 71.3  \\
GradCAM++~\citep{ChattopadhyaySH18wacvgradcampp} {\small \emph{(wacv,2018)}}  &                     84.4 & 72.4  \\
Smooth-GradCAM++~\citep{omeiza2019corr} {\small \emph{(corr,2019)}}  &                     82.6 & 75.2  \\
XGradCAM~\citep{fu2020axiom} {\small \emph{(bmvc,2020)}}  &                     87.3 & 71.6  \\
% LayerCAM~\citep{JiangZHCW21layercam} {\small \emph{(ieee,2021)}} &  &                     84.4 && 72.1  \\
TCAM~\cite{tcamsbelharbi2023} {\small \emph{(wacv,2023)}}  &                     84.4 & 72.1  \\
% CoLo-CAM~\cite{belharbi2023colo} {\small \emph{(corr,2023)}} &  &                     84.4 && 72.1  \\
TrCAM-V (ours) &                      \textbf{92.2} & \textbf{87.9}  \\
\bottomrule \\
\end{tabular}
}
\caption{Classification accuracy (\cl) on \ytovone~\cite{prest2012learning} and \ytovtwodtwo~\cite{KalogeitonFS16} test sets. Baseline results are borrowed from \cite{belharbi2023colo}.
}
\label{tab:cl-perf}
\vspace{-1em}
\end{table}
}

\noindent \textbf{Baseline Models.} To validate our TrCAM-V method, its performance is compared with various state-of-the-art methods, e.g., FPPVOS~\citep{umer2021efficient}, SPFTN~\citep{zhang2020spftn} POD~\citep{jun2016pod}, and~\citep{Croitoru2019, Halle2017, joulin2014, Kwak2015, papazoglou2013fast, prest2012learning, RochanRBW16, Tokmakov2016, tsai2016semantic}. The performance of our model is also compared with different CAM methods:  LayerCAM~\citep{JiangZHCW21layercam},  XGradCAM~\citep{fu2020axiom}, GradCAM++~\citep{omeiza2019corr}, GradCAM~\citep{SelvarajuCDVPB17iccvgradcam} and CAM~\citep{zhou2016learning}. Note that our method for generating pseudo-label and pseudo-pixel selection is generic, and can be integrated to train any model for WSOL.

%%%
\subsection{Comparison with State-of-Art Methods}

\noindent \textbf{Qualitative Results.} The classification and localization performance of our model is presented in Table \ref{tab:cl-perf} and \ref{tab:loc}, respectively. Regarding classification performance, our model exhibits an improvement of 4.9\% and 12.7\% on the \ytovone and \ytovtwodtwo datasets, respectively, when compared with the baseline model. Similarly, our model is able to surpass baseline models in terms of localization performance on both datasets. In contrast to the baseline methods, our method is able to achieve state-of-the-art performance without the need for exploiting temporal information, thereby enabling a consistent pipeline at both training and inference time.  In addition to this, we trained our model using pseudo-label harvested from \clipseg \cite{luddecke2022image} that helps in improving the performance of our model on \ytovtwodtwo dataset as shown in Table \ref{tab:loc}.

\noindent \textbf{Quantitative Results.} Fig.\ref{fig:results_yt_1} depicts localization predictions of TrCAM-V compared to baselines and state-of-the-art methods. Results illustrate that our model can generate robust maps that encompass both FG and BG regions, demarcated by sharp boundaries. Activation-based methods focus on discriminative areas that are common among different instances of the same class. Moreover, TCAM generates localization maps with sharp boundaries over various object parts, facilitating the prediction of bounding boxes that fully enclose the object. However, activation maps fail to encompass all parts of an object, resulting in inaccurate localization in produced maps.

%%%%
\subsection{Ablation Studies}
Ablations presented in Table \ref{tab:ablation-parts} are conducted to show the effectiveness of various loss terms. This study indicates that all auxiliary terms of our loss contribute significantly to the model's performance on both datasets, \ytovone and \ytovtwodtwo. Harvesting pseudo-labels from the \clip model yields competitive performance compared to the baseline method. Adding size terms to the loss function helps in connecting different object parts. Additionally, CRF terms help to significantly improve localization performance by aligning the boundaries of the localization map with the boundaries of the concerned object. In contrast to the baseline methods, TrCAM-V can achieve state-of-the-art performance without exploiting temporal information. 

{
\setlength{\tabcolsep}{3pt}
\renewcommand{\arraystretch}{1.1}
\begin{table}[ht!]
\centering
\resizebox{0.88\linewidth}{!}{%
\centering
\small
\begin{tabular}{lggcc}
% \multicolumn{2}{l}{\textbf{}} &  \multicolumn{3}{c}{\corloc} \\
\toprule
&  \multicolumn{2}{c}{\textbf{\corloc} } \\
\hline
Methods &  \ytovone & 
\ytovtwodtwo 
\\
% \cline{1-1}\cline{3-3}\cline{5-5}\\
\hline
% \multicolumn{3}{l}{Layer-CAM~\citep{JiangZHCW21layercam} {\small \emph{(ieee,2021)}}} &  &                     \tableplus{65.6}  \\
TCAM~\cite{tcamsbelharbi2023} {\small \emph{(wacv,2023)}}~\citep{JiangZHCW21layercam} & 73.0 & 72.2  \\
\hline
% \cline{1-4} \\
% \multirow{3}{*}{$n=0$} & 
% &       
PAL & 74.8 & 72.3       \\
% & & 
PAL + ASL &   75.5 & 74.2      \\
PAL + CRF &   81.4 & 76.2       \\
% && 
% && 
PAL + CRF + ASL &   84.8 & 76.7\\
\hline 
% $n>0$ & &
% Ours + PAL + CRF + ASC + CAM-TMP &  &   $73.0$ \\
% \cline{1-1}\cline{3-3} \\
Improvement &                           \tableplus{11.8} & \tableplus{4.5}\\
\bottomrule
\end{tabular}
}
\caption{\corloc accuracy of TrCAM-V with various losses.}
\label{tab:ablation-parts}
\vspace{-1em}
\end{table}
}

%%%%%%%%%%%%%%%%%%%%%%%%%
\subsection{Conclusion}
In this paper, a novel approach is proposed to train transformers for WSVOL tasks. The transformer consists of two heads for classification and localization, which are trained using video-level class labels and pixel-wise pseudo-labels, respectively. These pseudo-labels are harvested from the standard \clip model. Additionally, a CRF loss is employed to align the boundaries of the localization map and the concerned object. Our model outperforms state-of-the-art methods in terms of both classification and localization accuracy without requiring temporal information across subsequent frames. 

{
    \small
    \bibliographystyle{ieeenat_fullname}
    \bibliography{cas-refs}
}
% WARNING: do not forget to delete the supplementary pages from your submission 
% \input{sec/X_suppl}

\end{document}